\documentclass{article}

\usepackage{microtype}
\usepackage{graphicx}
\usepackage{subfigure}
\usepackage{booktabs} 
\usepackage{algorithm}
\usepackage{amsmath}
\usepackage{amssymb}
\usepackage{mathtools}

\usepackage{hyperref}

\usepackage[accepted]{icml2018}

\icmltitlerunning{Realistic Physics Based Character Controller}

\begin{document}

\twocolumn[
\icmltitle{Realistic Physics Based Character Controller}




\begin{icmlauthorlist}

\icmlauthor{Joe Booth}{vidya}
\icmlauthor{Vladimir Ivanov}{stanford}
\end{icmlauthorlist}

\icmlaffiliation{stanford}{Stanford University}
\icmlaffiliation{vidya}{Vidya Gamer, LLC}

\icmlcorrespondingauthor{Joe Booth}{joe@joebooth.com}
\icmlcorrespondingauthor{Vladimir Ivanov}{vivan@stanford.edu}

\icmlkeywords{Machine Learning, Reinforcemnet Learning, Animation, Simulation, Character Control}

\vskip 0.3in
]



\printAffiliationsAndNotice{}  

\begin{abstract}
Over the course of the last several years there was a strong interest in application of modern optimal control techniques to the field of character animation. This interest was fueled by introduction of efficient learning based algorithms for policy optimization, growth in computation power, and game engine improvements. It was shown that it is possible to generate natural looking control of a character by using two ingredients. First, the simulated agent must adhere to a motion capture dataset. And second, the character aims to track the control input from the user. The paper aims at closing the gap between the researchers and users by introducing an open source implementation of physics based character control in Unity framework that has a low entry barrier and a steep learning curve. 
\end{abstract}

\section{Introduction}
\label{intro}
Game engines have long been supporting physics based simulation of rigid body collisions that allowed to build natural looking virtual worlds that, with a certain amount of imagination allowed the user to dive into the experience. Nowadays, the cloth and fluid simulation have become ubiquitous. The adoption of physics based character control is not that widespread.

Machine learning is experiencing an unprecedented growth these years. Neural networks, being universal funtion approximators have shown their ability to model complex scenarios in Online Sales, Finance, and Computer Vision. Building animator controller requires much manual effort. They rely on graphs that involve multiple nodes and non trivial logic. Thus, they are perfectly suited for being modeled with neural networks.
There exist successful applications of Machine Learning techniques to animation. \cite{DBLP:journals/tog/HoldenKS17} uses neural networks to generate a natural looking kinematic controller with neural networks. The game developer community is interested in working with a robust controller that exhibits realistic behavior and interacts with a simulated environment. This is where the interests of the two communities, game developers and optimal control researchers, meet.

\begin{figure}[h!]
\vskip 0.2in
\includegraphics[width=1.\columnwidth]{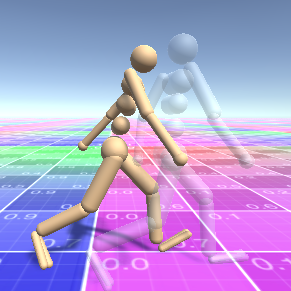}
\caption{ The agent and reference motion }
\label{fig:referenceMotion}
\vskip -0.1in
\end{figure}

\section{Related work}

The policy based techniques took off when it was shown that PPO \cite{DBLP:journals/corr/SchulmanWDRK17} algorithm manages to train a physics based Mujoco \cite{DBLP:conf/iros/TodorovET12} humanoid character to run by only using the joint rotations and positions, without prior knowledge of inverse kinematics, physics, or humanoid model. However, the moves that the agent learned were unnatural. The humanoid model doesn’t precisely repeat every bone and joint of a real human, and thus the optimal control learned isn’t guaranteed to look anywhere near to realistic. 

On the other hand, (Clavet, 2016) focuses on motion matching that focused on realistic behavior rather than the physics based interaction. Having a description of current character joint positions and velocities, and desired trajectory for the next several seconds obtained from the user input, they search the motion capture dataset for most similar looking motion and blend it with the current character motion. 

The DeepMimic authors \cite{DBLP:journals/tog/PengALP18} bridged the gap between the two approaches and introduced the technique of simultaneous motion capture tracking and physics based simulation. The authors of \cite{DBLP:journals/tog/BergaminCHF19} made the next step by introducing the user input into play, and use the PD controller as in \cite{DBLP:conf/mig/Chentanez0MMJ18}. 

\subsection{Existing Implementations}
The PPO algorithm, along with an array of other learning based algorithms are implemented in OpenAI Gym \cite{DBLP:journals/corr/BrockmanCPSSTZ16}. The DeepMimic \cite{DBLP:journals/tog/PengALP18} authors based their research on the Bullet physics engine \cite{DBLP:conf/siggraph/Coumans15} and open sourced their project. To our knowledge, there is no open source implementation of \cite{DBLP:journals/tog/BergaminCHF19}. Our work is based on ML Agents framework \cite{DBLP:journals/corr/abs-1809-02627} and extends the solution   \cite{DBLP:journals/corr/abs-1902-09097} that is similar to OpenAI Gym. Our code, videos and getting started tutorials are hosted at \url{https://github.com/Unity-Technologies/marathon-envs}.

\section{Experiments}

\subsection{Overview}
In this paper, we focus primarily on two tasks. First is mimicking animation. And second is following user input while maintaining similarity to an animation. Rather than using a motion capture dataset, we build upon standard Unity animation approach: an animation .fbx file is used as a source for an animator controller. We use a humanoid character for the analysis. 

\subsection{Character Controller with no User Input}
For the task of animation mimicking without following user control input, we use the phase of animation in the observation state. For looping animations, e.g. the Walking or Running, the phase periodically goes from 0 to 1, while for non-looping animations, e.g. Backflip, or Kick animation, the phase goes from 0 to 1 once. 

The joint rotations, positions, velocities and angular velocities are used as input as well. The features r of mass velocity and its position are added to the observation state as well. We also explore using the angular momentum for more complex tasks like Backflip.

For this task we use Unity’s ConfigurableJoints. The actions are used as target rotation for the joints. As a result, the model has observation space size of 258, and action space dimension of 21.

\begin{figure}[h!]
\vskip 0.2in
\includegraphics[width=1.\columnwidth]{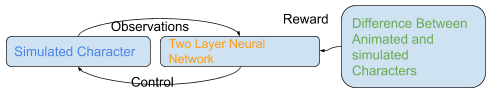}
\label{fig:envs}
\vskip -0.1in
\end{figure}

\subsection{Character Controller with User Input}
In this setup, we have an animated character that is controlled by user input. At the same time, we spawn another character, a physics based one, that aims to mimic the animated character.
As in the previous task, we use the physics based character’s joint rotation, positions, angular velocities and velocities to describe the observation features. However, here we also add the difference between the animated character’s joint features, and the physics based character features.
The phase input isn’t necessary for the user input task, since in this task we explicitly provide the animated character’s joint details to the model. We also add the information about the humanoid as a whole by providing the center of mass coordinates and velocity. 

For this task, we keep only a fraction of body parts for tracking. Therefore, the observation state size is smaller, 115. 

We also upgraded to the recently introduced ArticulationBody entity when working with joints. The actions are mapped to target positions of the joints. The ArticulatinoBody has an option of specifying stiffness and damping in addition to the target positions. We explore mapping actions to these parameters as well. The action dimension is at least 21 for the experiments.

\begin{figure}[h!]
\vskip 0.2in
\includegraphics[width=1.\columnwidth]{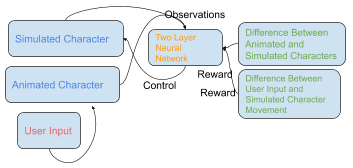}
\label{fig:envs}
\vskip -0.1in
\end{figure}

\section{Results}
\subsection{Character Controller with no User Input}

We aim at delivering a feasible behavior within 128 million learning steps. Which corresponds to approximately 24 hours’ training on an average desktop machine. Using GPU as a tensorflow backend only reduces the training speed in our experience. The reward terms weights are kept unchanged over the span of all animations we trained. We also use early stopping condition: when reward falls below a distinct for each animation threshold, we interrupt training. Setting correct threshold is crucial, for example, to prevent a humanoid that is supposed to be running from running on knees.

\begin{figure}[h!]
\vskip 0.2in
\includegraphics[width=1.\columnwidth]{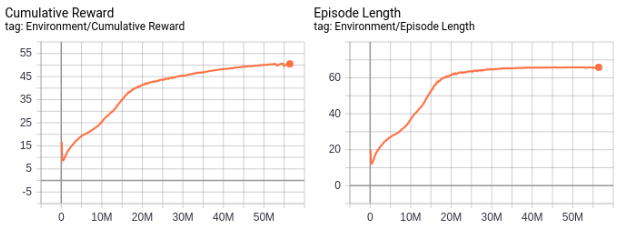}
\caption{A typical learning process is characterized by steadily increasing reward and episode length. The learning curves for learning Kicking animation are displayed on the figure. \textbf{Left:} Cumulative reward against the number of training steps. \textbf{Right:} Episode length against number of learning steps}
\label{fig:envs}
\vskip -0.1in
\end{figure}

\begin{figure}[h!]
\vskip 0.2in
\includegraphics[width=1.\columnwidth]{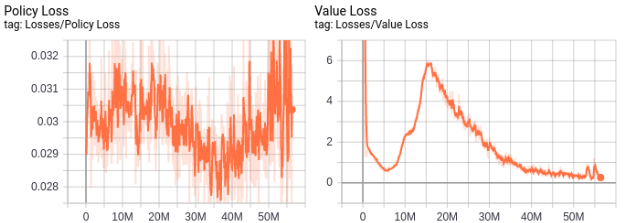}
\caption{We use an actor-critic model with PPO algorithm. Value loss function first grows, and decreases only after the reward plateaus \textbf{Left:} Policy loss against number of training steps \textbf{Right:} Value loss against number of training steps}
\label{fig:envs}
\vskip -0.1in
\end{figure}

\subsection{Angular Momentum in Observations}
\begin{figure}[h!]
\vskip 0.2in
\includegraphics[width=1.\columnwidth]{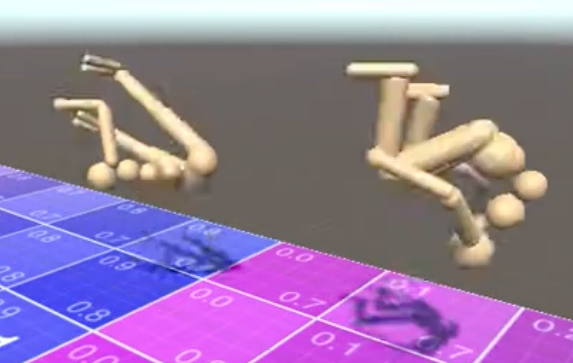}
\caption{ Backflip trained with no angular momentum term \textbf{Left:} Simulated agent. Note the straight legs and tilt \textbf{Right:} Reference motion}
\label{fig:backflip}
\vskip -0.1in
\end{figure}

The authors of DeepMimic paper mentioned a peculiarity of backflip training. The agent would never learn to make a full mid air flip. It would make a jump and lift its leg only. We also observed this event when training. The DeepMimic paper treats this by initializing the agent at a sampled state from reference motion. However, we found that we can do without it. Our trick is to add angular momentum to observations. And use the difference between the angular momentum of reference motion and the simulated character’s angular momentum as a reward signal. When a motion is mostly based on balancing, which is the case for Walking and Running animations, the term isn’t necessary. For backflip, however, the rotational part of the motion cannot be neglected. Contrast the two learned behaviors with and without the angular momentum term shown on Figure~\ref{fig:backflip}. 

\subsection{Character Controller with User Input}
To benchmark the performance of our implementation with the presence of user control, we set up training with random control input that picks a direction from a range of [-45, +45] degrees. The power of input is sampled from a range of [0, 1], where 0 corresponds to Idle animation and 1 to Running. 

\begin{figure}[h!]
\vskip 0.2in
\includegraphics[width=1.\columnwidth]{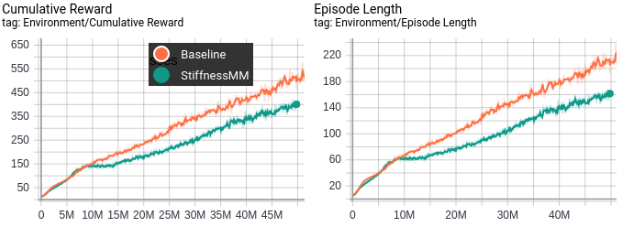}
\caption{ Compare the performance of a learnt multiplier and constant multiplier, the baseline }
\label{fig:stiffnessMM}
\vskip -0.1in
\end{figure}

The work \cite{DBLP:journals/tog/BergaminCHF19} uses a PD controller to filter the predicted actions of the neural network. The Unity's ArticulationBody joints we use in our work provide a possibility of specifying a target of rotation, along with stiffness and damping. This is similar to the behavior of a PD controller. We benchmarked the performance of the agent when the damping and stiffness parameters are learnt. We initialize stiffness at 30, and damping at 100. If the neural network learns multiplier of the initial stiffness and damping, the performance gets worse: see Figure~\ref{fig:stiffnessMM}. The reasoning is probably that the multiplier provides too steep changes to the agent. 

\begin{figure}[h!]
\vskip 0.2in
\includegraphics[width=1.\columnwidth]{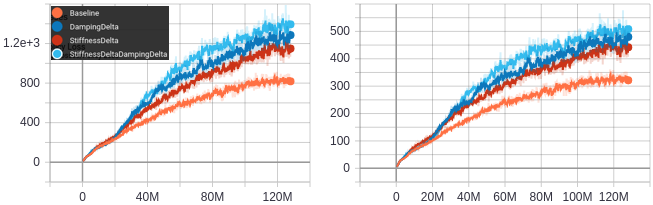}
\caption{ Here we run 4 separate experiments. First, we fix the damping and stiffness. Next, we unfreeze stiffness. In the third one, we freeze stiffness and learn damping addition. Best results are achieved when both damping and stiffness additions are learnt simultaneously }
\label{fig:dampingDeltaStiffnessDeltaLearnt}
\vskip -0.1in
\end{figure}

On the other hand, When the agent learns an addition to the initial damping and stiffness, the reward gets better. See Figure ~\ref{fig:dampingDeltaStiffnessDeltaLearnt}.

\subsection{Reference Motion teleport}
\begin{figure}[h!]
\vskip 0.2in
\includegraphics[width=1.\columnwidth]{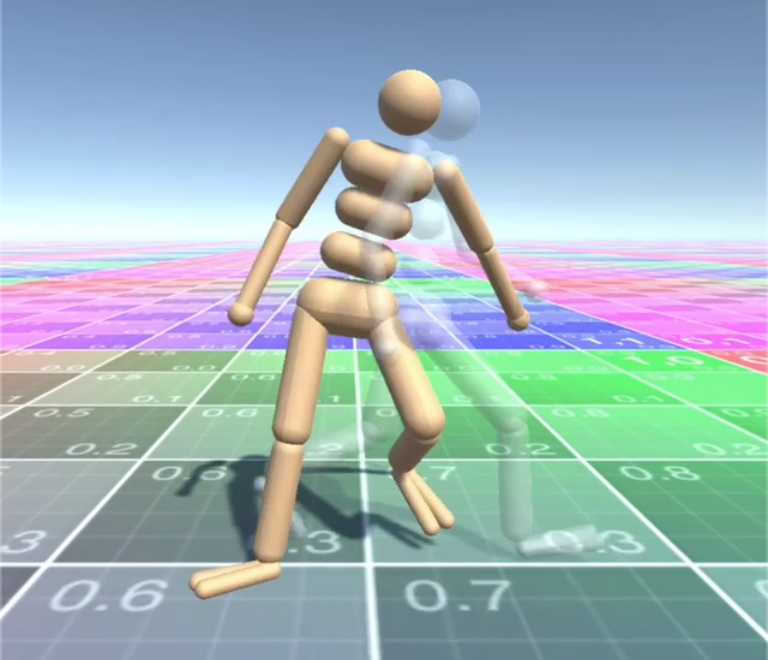}
\caption{ Humanoid learns a 180 degree turn while keeping balance }
\label{fig:turn}
\vskip -0.1in
\end{figure}

In the previous experiments we reset the scene once the reward falls below a predefined threshold. The controller direction is sampled from a fixed range of values. In this experiment, we reset the scene only when the agent falls on ground. And the direction range isn't capped. In case the reward drops lower than a predefined threshold, we teleport the reference motion to the center of mass coordinates of the simulated agent. As a result, the agent learns to make a 180 degree turn, even though the programmed animator controller we use for reference motion doesn't include the animation for the turn. This is shown on Figure~\ref{fig:turn}

\section{Conclusion}

In this paper we present an implementation of realistic Physics based character control. The implementation provides an easy to learn framework for a widely used game engine. Game developers, optimal control researchers and enthusiasts are welcome to introduce new animations, and train agents within a reasonable time frame. 

\section{Future work}
For the future work a number of improvements can be made. This work presents benchmarks for a simple controller. A more sophisticated controller and new moves beyond Walk and Run can be introduced. For example, Backflip and Jump. The reference motions we work with are animations, and we employ animator controllers. An important addition would be using motion capture data that would enrich the behavior of the agent. This can be done using Kinematica package. Right now, even though the agent looks natural, and it is able to closely track the reference motion, they often do not match closely, see Figure~\ref{fig:referenceMotion}. Training with a motion capture dataset, or with a non-trivial animator controller can lead to improvements. 


\bibliography{example_paper}
\bibliographystyle{icml2018}

\end{document}